\documentclass[fleqn,10pt]{wlscirep}
\usepackage[utf8]{inputenc}
\usepackage[T1]{fontenc}

\usepackage{algorithm}
\usepackage{algorithmic}
\usepackage{amsfonts}
\usepackage{amsmath}
\usepackage{multirow}
\usepackage{array}
\usepackage{booktabs}
\usepackage{caption}
\usepackage{xcolor}
\usepackage[capitalize,noabbrev]{cleveref}
\usepackage{newfloat}
\usepackage{listings}
\usepackage{url}

\DeclareMathOperator*{\argmax}{arg\,max}

\newtheorem{definition}{Definition}
\newtheorem{remark}{Remark}

\newcommand{\blue}[1]{{\color{blue}#1}}

\title{Hierarchical Meta-Reinforcement Learning \\ via Automated Macro-Action Discovery}

\author[1, *]{Minjae Cho}
\author[2, +]{Chuangchuang Sun}

\affil[1]{University of Illinois Urbana-Champaign, Department of Aerospace Engineering, Urbana, 61801, USA}
\affil[2]{Mississippi State University, Department of Aerospace Engineering, Starkville, 39762, USA}

\affil[*]{minjae5@illinois.edu}
\affil[+]{csun@ae.msstate.edu}


\begin{abstract}
Meta-Reinforcement Learning (Meta-RL) enables fast adaptation to new testing tasks. Despite recent advancements, it is still challenging to learn performant policies across multiple complex and high-dimensional tasks. To address this, we propose a novel architecture with three hierarchical levels for 1) learning task representations, 2) discovering task-agnostic macro-actions in an automated manner, and 3) learning primitive actions. The macro-action can guide the low-level primitive policy learning to more efficiently transition to goal states. This can address the issue that the policy may forget previously learned behavior while learning new, conflicting tasks. Moreover, the task-agnostic nature of the macro-actions is enabled by removing task-specific components from the state space. Hence, this makes them amenable to re-composition across different tasks and leads to promising fast adaptation to new tasks.
Also, the prospective instability from the tri-level hierarchies is effectively mitigated by our innovative, independently tailored training schemes. Experiments in the MetaWorld framework demonstrate the improved sample efficiency and success rate of our approach compared to previous state-of-the-art methods.
\end{abstract}

\begin{document}

\flushbottom
\maketitle
%
%

\section{Introduction} \label{sec:introduction}
Reinforcement Learning (RL) has rapidly advanced, enhancing decision-making capabilities in high-dimensional and complex systems. Recent innovations have expanded RL to address diverse challenges, including multi-task RL, safe RL, and offline RL, thus broadening its practical applications. Among these advancements, meta-learning emphasizes training an agent to adapt to new tasks by leveraging previously acquired knowledge. This often involves balancing the performance across multiple training tasks, exposing the learner to diverse task-parameters to facilitate future adaptation to any prospective test task within the same distribution.

Hence, due to the inherent challenge in meta-learning \cite{hospedales2021meta}, common methodologies are typically tested on narrow task distributional learning scenarios such as a jumping robot with varying loads (altering transition dynamics) and a running agent with different velocity criteria (changing optimality). Although it faces challenges when extended to more complex meta-learning scenarios, task representation learning \cite{hausman2018learning} shows its potential in such environments by identifying shared structures across given training tasks. The discovered representations offer fundamental insights into tasks, aiding the decision-maker in adapting and performing effectively on new tasks.

\begin{figure}[t]
    \centering
    \includegraphics[width=0.475\textwidth]{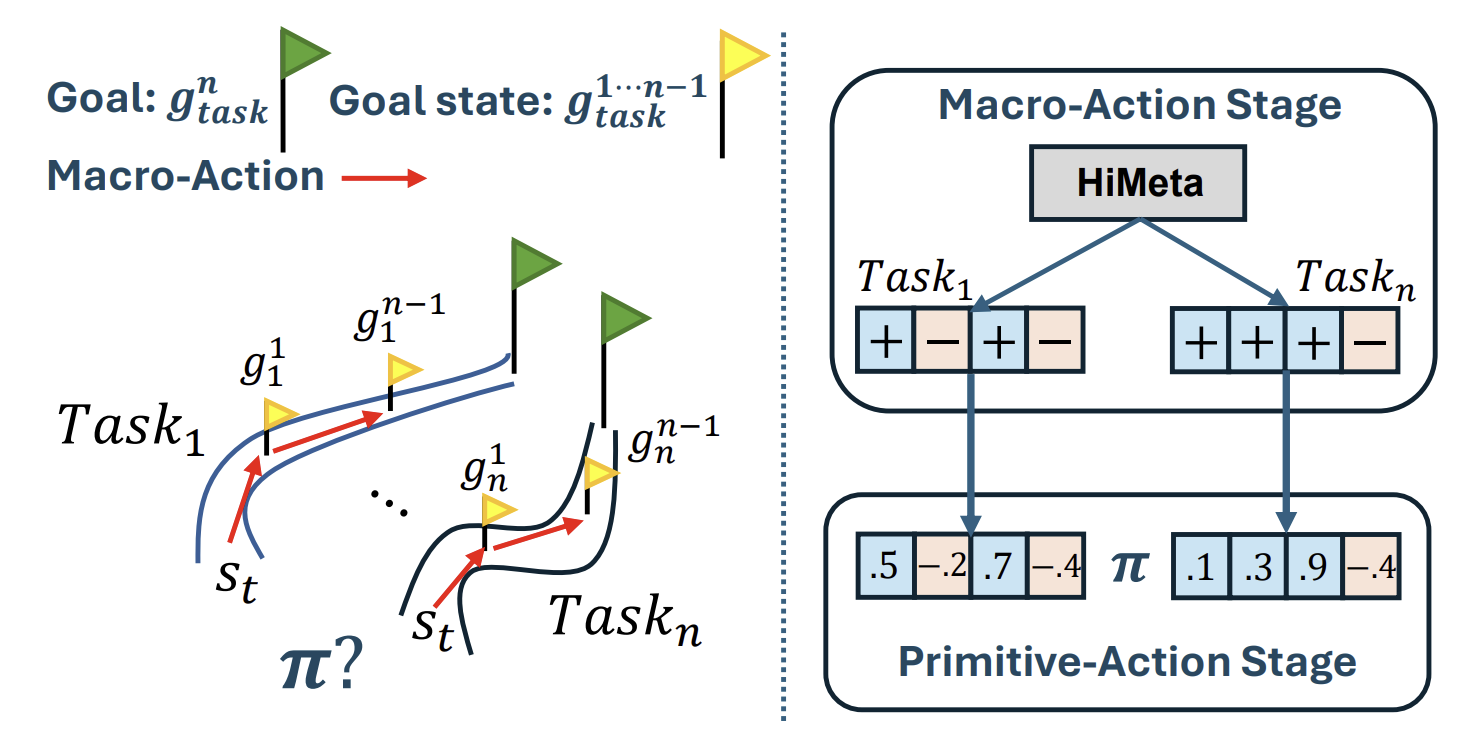}
    
    \caption{This overview outlines our approach. Our higher-level architecture, HiMeta, offers high-level directional action predictions to the policy, on which the policy generates primitive actions based on these directions. Additionally, the directional action prediction is trained using a modified VAE designed to link the current state to the desired goal state. This enables the effective distribution of decision-making tasks, facilitating solutions for complex, high-dimensional multi-task and meta-learning scenarios.}
\label{fig:himeta_illust}
\end{figure}

Building on task representations, which expand the input space with additional conditioners, we aim to further distribute the policy's information load by utilizing high-level actions, or \emph{macro-actions}. Specifically, macro-action provides directional action vectors to the low-level policy. Then, the policy fills with detailed low-level control input along the given direction. This approach is particularly useful for learning multiple tasks and adapting to new ones in hard-parameter sharing. In this setting, effective distribution of information processing can handle challenging task-specific instructions, ensuring that the policy retains its previously acquired optimal behavior in a particular task while learning new tasks. The illustration is given in \Cref{fig:himeta_illust}.

In short, beyond the existing bi-level representation learning (task-representation | low-level control), we introduce a tri-level method (task-representation | high-level control | low-level control) that includes an additional layer for constructing macro-actions. For task-representation, we adopt the goal state concept from hierarchical RL literature \cite{nachum2018data, dwiel2019hierarchical}. For high-level control inputs, we propose a modified Variational AutoEncoder (VAE) to identify an effective placeholder for low-level control. This placeholder provides both sign-based and magnitude-level guidance, enabling the policy to refine it into precise low-level controls. In the architecture, the encoder predicts macro-actions, while the decoder uses imputations to bridge the current state with the goal state. Therefore, this approach brings appealing properties in complex, sparse, long-horizon problems. We provide mathematical views of the mechanism of the encoder and decoder.

For stacking three levels of layers, we design each to have an independent role in decomposition to avoid the ``curse of hierarchy" \cite{harutyunyan2019hindsight, levy2017learning}, which refers to the instability of learning higher-level layers caused by frequent dynamical transitions by the low-level decision-maker. While this challenge has previously limited hierarchical learning to bi-level layers, our novel solution of independent training schemes tailored for each hierarchical layer effectively enables three-layer hierarchical learning, demonstrating its effectiveness in broad meta-task learning. Our contributions are as follows:
\begin{itemize}
    \item Proposing a novel architecture to automatically discover \emph{macro-actions} on which the policy fills the low-level input. This additionally prevents the loss of optimality in a particular task while learning news by the distribution of information load.
    \item Introducing a training scheme and techniques to avoid the ``curse of hierarchy,'' enabling multi-level layers and collaborative learning across multiple tasks.
    \item Conducting extensive meta-learning experiments on complex and diverse task distributions, achieving state-of-the-art performance compared to baselines.
\end{itemize}

This letter outlines the preliminaries of meta-learning, offering insights into hierarchical decomposition in the following section \cref{section:Preliminaries}. Additionally, we detail our algorithmic approach and highlight the key insights that drive the state-of-the-art performance of our method in \cref{section:HiMeta}, accompanied by extensive experiments in \Cref{section:Experiment}. Finally, we provide our concluding remarks and discuss potential improvements and future work in \cref{section:concluding remarks}.

\section{Preliminaries} \label{section:Preliminaries}
Reinforcement learning operates within the mathematical framework of Markov Decision Processes (MDP) \cite{puterman2014markov}, characterized by the tuple \(\langle\mathcal{S}, \mathcal{A}, T(\cdot), R(\cdot), \gamma \rangle\), where \(\mathcal{S} \in \mathbb{R}^s\) represents the state set (observation), \(\mathcal{A} \in \mathbb{R}^a\) is the action set, \(T(\cdot): \mathcal{S} \times \mathcal{A} \rightarrow \mathcal{S} \) denotes the transition function, and \(R(\cdot): \mathcal{S} \times \mathcal{A} \rightarrow \mathbb{R}\) is a reward function with \(\gamma\) as a discount factor. RL for any arbitrary task, $\mathcal{T}$, comprises its own transitional and reward dynamics. Formally, the definition of a task, \(\mathcal{T}\), is \(\langle T(\cdot), R(\cdot) \rangle \in \mathcal{T}\).

The goal of deep RL is to find a policy parameterized by \(\theta\), denoted as \(\pi_\theta(a|s)\), that maps the state to an optimal action, yielding the highest cumulative rewards (optimal trajectory) under task $\mathcal{T}$, as shown below:

\begin{equation}
    \label{eqn:mtl-training-scheme}
    \max_{\theta} \mathcal{J}(\pi_\theta, \mathcal{T}) = \mathbb{E}_{\substack{a \sim \pi_\theta(s),\\ s' \sim T(s,a)}} \left[ \sum_{t=0}^T \gamma^t R(s_t, a_t) \right] 
\end{equation}

\subsection*{Meta-Learning}
Meta-learning extends multi-task learning by aiming for rapid adaptation to unseen tasks using prior knowledge. Formally, the trained policy aims to maximize the prospective optimality objective of the test task $\mathcal{T}_{\text{test}}$:
\begin{equation}
    \begin{aligned}
        \max_\theta \mathcal{J}(\pi_\theta, \mathcal{T}_{\text{test}}), \quad \text{where} \quad \theta = \theta - \nabla \mathcal{J}(\pi_\theta, \mathcal{T}_{train})
    \end{aligned}
\end{equation}
Meta-training is typically achieved by conditioning task representations or finding a parametric space that is close to the new task's optimal space as above.

Task representation aims to discover common or distinct knowledge across tasks to better analyze the test tasks. This is typically done by posterior sampling over the variable of interest $x$, a latent task representation, using a variational autoencoder with the following objective:

\begin{equation}
    \log p(x) = \mathbb{E}_{q(z|x)}[\log p(x|z)] - \text{KL}(q(z|x) \| p(z))
\end{equation}
where \(p(\cdot)\) denotes the probability, \(q(\cdot)\) represents the posterior distribution, and KL is Kullback-Leibler divergence. The representation is a type of informative label, and it is then conditioned to the policy, $\pi_\theta$, to endow better differentiation. This creates useful input spaces that are easily shared across multiple tasks, inducing positive knowledge sharing. The construction of posterior distribution of representations has proven effective in many previous studies involving complex meta-adaptation tasks, which are relevant to our approach. Details of relevant methods are described in Related Works.

\subsection*{Hierarchical Reinforcement Learning}
Hierarchical RL is introduced as a solution for tasks with long horizons and sparse rewards by decomposing the original problem into several sub-problems. This is typically achieved with a new optimality objective, \(\hat{\mathcal{J}}\), as shown below by additionally conditioning the policy on a \emph{goal} \(z\):

\begin{equation}
    \hat{\mathcal{J}} = \mathbb{E}_{\substack{a \sim \pi_\theta(s,z), \\ s' \sim T(s,a)} }\left[ \sum_{t=0}^T \gamma^t R(s_t, a_t) + \alpha \mathbb{I}(s_t = z) \right]
\end{equation}
where \(\mathbb{I}(\cdot)\) is an indicator function that adds a reward bonus, scaled by \(\alpha\), when the policy successfully reaches the given goal \(z\). The \(z\) can be a latent representation, \emph{goal}, or a raw state, \emph{goal state}. With this definition, the policy is efficiently guided in sparse rewards and long horizons settings \cite{dwiel2019hierarchical, hutsebaut2022hierarchical, harutyunyan2019hindsight, zhu2022hierarchical}.

\section{HiMeta: \\Hierarchical Meta-Reinforcement Learning} \label{section:HiMeta}
A high information load on the decision-maker can hinder learning by overwhelming its capacity to process information. This is evident in multiple-task settings where learning one task can downgrade the behavior in another. Thus, we propose decomposing representations into a more fundamental, task-agnostic language for an ``action-guided" learning approach. We introduce three handcrafted definitions of goal states and add a layer that finds the future action embedding that can connect the current state to the goal state. In short, we propose carefully designed architectures and loss functions that discover macro-actions by bridging the action space to the state space using VAE and imputations.

\begin{figure}[t]
    \centering
    \includegraphics[width=0.475\textwidth]{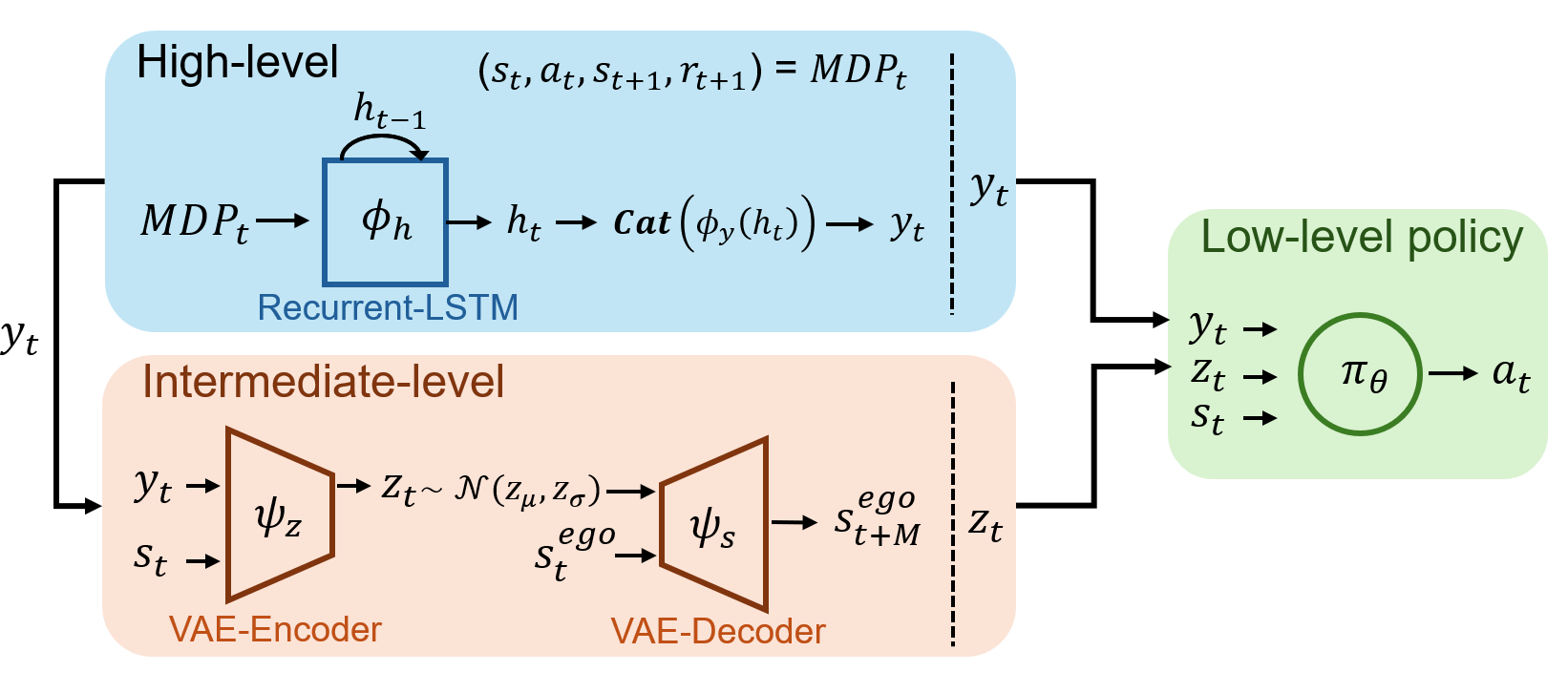}
    
    \caption{Our algorithm consists of three hierarchical layers: high (task representation learning), intermediate (macro-action discovery), and low (primitive actions discovery). The high-level layer discovers a task representation \(y\), using a recurrent unit, simultaneously with value-function training. The intermediate layer analyzes the representation given the current state to determine the macro-action \(z\) which is the (+/-) sign of actions. This is trained via VAE using the missing information technique. The state element \(s^{ego}\) is a subset of the state set, $\mathcal{S}^{ego} \in \mathcal{S}$ that includes everything except the agent's self-state. The subsequent loss in decoder with \(s^{ego}\) will shape macro-actions, \(z\), task-agnostic and compact. The representation and macro-action are then conditioned in the low-level policy to make primitive decisions. Gradients do not overflow between hierarchical layers, ensuring each layer's independent role.}
\label{fig:himeta_schematic}
\end{figure}

\subsection*{High-level layer}
The High-Level (HL) architecture aims to discover latent task representations as a probability distribution that reconciles the trajectory data across multiple tasks. Our HL module utilizes recurrent units for modeling the probability distributions. We use a categorical feed-forward network with an information-theoretic regularizer \cite{lee2023parameterizing} to induce explorations while penalizing allocations if it is not worthwhile.

To align our experimental settings with the baselines in the Experiment section, we use a GRU \cite{cho2014learningphraserepresentationsusing} as the recurrent unit, though an LSTM \cite{article} can also be an alternative. For simplicity, we denote the HL layer's parameters by \(\phi\) and use subscripts to indicate specific sub-architectures. For example, the recurrent model in \Cref{fig:himeta_schematic} is represented as \(\phi_h\), where \(\phi_h\) denotes the encoder giving output \(h\). The HL feed-forward process is denoted given $\text{MDP}_t = (s_t, a_t, r_t, s_{t+1})$:

\begin{equation}
    \begin{aligned}
    h_t =&\  \phi_h (\text{MDP}_t, h_{t-1}), \\
    y_t =&\  \text{Cat}(\phi_y(h_t))
    \end{aligned}
\end{equation}
where $h_{t-1} = \phi_h(\text{MDP}_{0:t-1})$ is the hidden state of the recurrent unit, $\phi_y$ is a feed forward neural network, and $\text{Cat}$ is a categorical activations in to discrete probability space.

\subsubsection*{HL-Training Scheme}
Our HL design includes an encoder where the representation is inferred while being simultaneously trained alongside a value function using the Mean Squared Error (MSE) loss:
\begin{equation} \label{eqn:value_loss}
    \mathcal{L}_{HL(V^{\pi})} = [V^{\pi_\theta}(\phi(MDP_t), s_t) - V(s_t)]^2 
\end{equation}
where the first term is our estimation with the learned representations, and the second is a true state value function. The true value function is estimated via Monte-Carlo samplings.

Inspired by \cite{lee2023parameterizing}, entropy regularizer is used to encourage exploration of representations within a task \(\mathcal{T}\), while occupancy loss ensures having likely correct task representations only when it successfully reduces the value loss. Formally, the entropy regularizer is defined as:
\begin{equation}
    \mathcal{L}_{HL(ent)} = \log \frac{p(y_t)}{\phi_y(y_t|h_t)}
\end{equation}
and the occupancy loss is:
\begin{equation}
    \begin{aligned}
        \mathcal{L}_{HL(occ)} = - (\mathbf{e}^T y_t)\log K 
    \end{aligned}
\end{equation}
where  \(K\) is the task inference dimension and $\mathbf{e} = [ e^{-K+1}, e^{-K+2}, \dots, e^{-1}, e^0 ]\in\mathbb{R}^K$. We scale it with the upper bound of the entropy regularizer \(\log K\), encouraging their competition.
Thus, by creating the HL objective with scaling parameters, the high-level architecture can successfully and efficiently infer representations $y_t$ for use by the lower-level architectures. The HL update is to optimize the following loss function \(\mathcal{L}_{HL}\) as
\begin{equation} \label{eqn:HL_loss}
    \begin{aligned}
        \min_{\phi}  \alpha_V \mathcal{L}_{HL(V^{\pi})}
        + \alpha_{\epsilon}\mathcal{L}_{HL(ent)} + \alpha_o\mathcal{L}_{HL(occ)}
    \end{aligned}
\end{equation}
The above training of the HL parameters $\phi$ is run across \(N\) tasks.

\subsection*{Intermediate-Level Layer}
Our key contribution is the Intermediate-Level (IL) architecture. At this level, we infer a macro-action using a modified VAE. The objective is to further distribute the decision burden from the policy and guide the policy to the desired state space using goal-conditioning. This is fundamentally different from previous hierarchical methods that only aimed to reach the goal state while neglecting the complexity of learning systems such as multi-task and meta-learning. 

Our design leverages a missing-information technique, formally imputations \cite{hong2023imputation, MCCOY2018141, peis2022missingdataimputationacquisition}, which uses relevant data to infer the expected data pattern in a missing region. Given two arbitrary points in the state space where the actions to transition from one to the other are unknown, we aim to discover macro actions, \(z\), that connect them. The term ``\emph{macro-}" is used as a temporally extended action:
\begin{definition}\label{def:macro-action}
    (Macro-action) Given a function \( f(\cdot) \), a macro-action, \( z \), is defined as information that connects the current state to the desired goal state within the state space: \( s_{t+M} = f(s_{t}, z) \) where \(M \in \mathbb{N}_+\).
\end{definition}

The overall feed-forward process for macro-action discovery by the encoder is as follows:
\begin{equation} \label{eqn:z-inference}
    z_t \sim \text{tanh}(\mathcal{N} (\psi_{z_\mu}(y_t, s_t), \psi_{z_\sigma}(y_t, s_t))),
\end{equation}
where a Multivariate Gaussian distribution $\mathcal{N}(\cdot, \cdot)$ is constructed with mean $\psi_{z_\mu}(\cdot, \cdot)$ and variance $\psi_{z_\sigma}(\cdot, \cdot)$.

\subsubsection*{IL-Training Scheme}
Imputation here is to find a proper action space embedding $\mathcal{Z} \approx \mathcal{A}$ that satisfies $\mathcal{S} \times \mathcal{Z}\rightarrow\mathcal{S}$. This is a complex problem, as the encoder and decoder have to concurrently bridge state and action spaces. Without proper treatment, \Cref{eqn:z-inference} will collapse, providing meaningless information only to lower the loss. Therefore, we modify the existing architecture and loss function of VAE to not only perform the standard VAE operations but also bridge the two spaces by replacing the prior distribution \( p(z) \) in the Evidence Lower Bound (ELBO) with action distributions. We also provide the state as input to the decoder to make it a transition predictor.

This strategy reshapes the latent space of the VAE by aligning the current belief, \( \psi_z(y_t, s_t) \), with the action space. In other words, the encoder, \( \mathcal{S} \times \mathcal{Y}\rightarrow \mathcal{A} \), remains within the action space given the state, while the decoder learns a function \( f(\cdot): \mathcal{S} \times \mathcal{A} \rightarrow \mathcal{S} \) in \Cref{def:macro-action} that maps the state and action to the future state. This models the decoder as a cumulative transition function since it predicts $M$ steps forward future state with current state and macro-actions. This shapes the macro-actions as a compressed action embeddings within the action space and confined in the domain of $[-1, 1] \sim \tanh(\cdot)$.

This represents our novel contribution, which learns the latent macro-action to directly assist the policy in reducing the decision-making challenge in complex multi-task execution using successful goal states from the policy. This prevents the loss of optimal behavior in some tasks while learning others. Additionally, we introduce the ego-state for task-agnostic inference by masking task-related state-wise elements:
\begin{definition} \label{def:ego-state}
    (Ego-state) Given a state set \(\mathcal{S}\), the set \(\mathcal{S}\) consists of two subsets: ego-state and other-state, such that \(\mathcal{S}_{ego} \oplus \mathcal{S}_{other} = \mathcal{S}\). The ego-state, \(\mathcal{S}_{ego}\), includes only self-state elements.
\end{definition}

\begin{remark}
    An example of an ego state is a blind robot. For the robot that is to perform any arbitrary task (e.g., navigating, grasping, or reaching), the task-agnostic components are only its body-related components: angle of joints and relative arm/body position. While state space decomposition is beneficial for task-agnostic learning \cite{hutsebaut2022hierarchical}, this instead voids out the necessary task information so that inference can stay in high-level space instead of detailed execution.
\end{remark}
The ego-state is particularly appealing for distributing decision-making into a set of independent roles. Specifically, macro-actions provide high-level action embedding, while the policy executes the detailed actions. Finally, the macro-action is then trained over the encoder parameters $\psi_z$ and decoder parameter $\psi_s$ by minimizing the following loss:
\begin{equation} \label{eqn:IL_loss}
    \begin{aligned}
        \min_{\psi_z, \psi_s} \text{ELBO}_t(y_t, z_t, s_t, s^{ego}_t, s^{ego}_{t+M}) \\
        \text{ELBO}_t(\cdot) = \ \mathcal{L}_{IL(KL)} + \mathcal{L}_{IL(trans)} \\
    \end{aligned}
\end{equation}
Given the decoder's forward pass 
$s_{t+M}^{ego} = \psi_s(y_t, z_t, s_t^{ego})$, each loss term is given below:
\begin{equation} \label{eqn:IL_specific_loss}
    \begin{aligned}[t]
        \mathcal{L}_{IL(KL)} = & \ \log \frac{p(a_t|y_t, z_t, s_t)}{\psi_z(z_t|y_t, s_t)} \\
        \mathcal{L}_{IL(trans)} = &\  \sum_{t=0}^T \log p (\psi_s(s^{ego}_{t+M}|y_t, z_t, s^{ego}_t)) \\
    \end{aligned}
\end{equation}
This approach is notably distinct from previous methods: SD \cite{lee2023parameterizing}, VariBad \cite{zintgraf2019varibad}, LDM \cite{lee2021improving}, $\text{RL}^2$ \cite{duan2016rl}, and PEARL \cite{rakelly2019efficient}, as it involves shaping the prior with the action space and incorporating additional ego-state input into the decoder to shape it as a transitional function.

In \Cref{eqn:IL_specific_loss}, \(s_{t+M}\) and \(s^{ego}\) are goal state and ego-state respectively. The goal state is the state the policy aims to achieve, typically at the designer's discretion, or it can be discovered using additional techniques. In our approach, we construct three definitions of goal state at the end of this section. 

\subsubsection*{Goal State Generation}
The goal state, \(s_{t+M}\), adapted per our method, is defined as follows:
\begin{enumerate}
    \item \textbf{Constant Discretization (CD)}: This naively discretizes the given episodic timelines into several segments. That is, we fix \(M\) as constant, thus obtaining a list of goal states: \([s_{M}, s_{2M}, \dots, s_{nM}]\).
    
    \item \textbf{Constant Margins (CM)}: This method maintains a constant margin between the current state and the goal state, resulting in a list of \([s_{M}, s_{1 + M} \dots s_{t + M}]\), given a constant \(M\).
    
    \item \textbf{Adaptive at Sub-Task (ST)}: {This uses the inferred sub-task from the HL layer, analyzing when the label changes in timescale. For example, given an array of states \(S = [s_0, s_1, \dots, s_t]\) and representations \(Y = [y_0, y_1, \dots, y_t]\), goal states are states where the representations change in time scale (i.e., \(\exists t: [y_0:y_{t-1}] \neq [y_1:y_t] \rightarrow S[t]\)).
    }
\end{enumerate}
As noted, goal states heavily rely on the policy's exploration. However, this lessens the information load by preventing the policy from forgetting previously learned tasks while it acquires new ones. In essence, although goal states are learned concurrently, this strategy ultimately helps the policy shift focus away from past tasks by relying on learned macro-actions for \emph{optimally} learned tasks.

\subsubsection*{Low-Level Layer}
For policy training, which is the objective of the Low-Level (LL) layer, we chose Proximal Policy Optimization (PPO), a practical solution that effectively reduces sample complexity using an off-policy gradient clipping approach. While there are abundant details about PPO, we briefly introduce its idea here, leaving parametric details in the Appendix. PPO maximizes the following objectives:
\begin{equation} \label{eqn:ppo_objective}
    \pi_\theta = \argmax_\theta \mathbb{E} \bigl[ \mathcal{J}(\theta, \mathcal{T}) - \alpha_1 \mathcal{L}(\theta, \mathcal{S})  + \alpha_2 \mathcal{E}(\theta) \bigr]
\end{equation}
where \(\mathcal{J}(\theta, \mathcal{T})\) is the gradient-clipped off-policy advantage for optimality, \(\mathcal{L}(\theta, \mathcal{S})\) is the value loss as in \Cref{eqn:value_loss}, \(\mathcal{E}(\theta)\) is the entropy loss for sufficient exploration, and \(\alpha_1\) and \(\alpha_2\) are scaling parameters. 

Note that our LL is trained without the value loss in \Cref{eqn:ppo_objective} since it is trained with the HL objective (\Cref{eqn:value_loss}). The feed-forward inference of the LL layer is:
\begin{equation}
    \begin{aligned}
        a_t = \pi_\theta(y_t, z_t, s_t)
    \end{aligned}
\end{equation}

For effective and efficient collaboration between the macro-action and primitive-action, we additionally construct an intrinsic reward function that adds a bonus to the policy when it follows the given macro-action \emph{sign-wise (+/-)}\footnote{This assumes the action space spans both negative and positive spaces. MetaWorld, our testbed, has the range $\mathcal{A} \in [-1, 1]$.}. The reason for the sign-wise bonus is to assign independent roles in decision-making to macro-actions and primitive actions as high-level and low-level actions respectively.

The optimality objective, \(\mathcal{J}(\theta, \mathcal{T})\), includes additional intrinsic reward incurred achieving macro-actions, starting with the non-clipped advantage function below.
\begin{equation}
    A(s,a) = \mathbf{r} + \gamma V(s') - V(s)
\end{equation}
where $\mathbf{r} = r_{in} + r_{ext}$ and $V(s) = \mathbf{r} + \gamma V(s')$

To carefully design an intrinsic reward signal that does not disrupt the original task optimality, it is scaled with an extrinsic reward signal. The one-time step intrinsic reward function is defined as:
\begin{equation}
    r_{in} = r_{ext} \cdot \frac{1}{|\mathcal{A}|} \sum_\mathcal{A} \mathbb{I}\big[ sgn(z) = sgn(a) \big]
\end{equation}
The intrinsic reward signal \( r_{in} \) is high only when the extrinsic reward signal is high and is bounded by the maximum value of the extrinsic reward. This enables the policy to maintain optimal behavior in tasks it has already mastered while still allowing for exploration in tasks it is still learning. Pseudo-code is given in \Cref{alg:1}.

\begin{algorithm}[t]
    \caption{HiMeta: Hierarchical Meta-RL}\label{alg:1}
    \begin{algorithmic}[1]
        \REQUIRE Network: $\phi$ (HL), $\psi$ (IL), $\theta_\pi$ (Policy), $\theta_V$ (Value)
        \REQUIRE Optimizers: Adam $\sim \phi, \psi, \theta$
        \REQUIRE Prepare data buffer $\mathcal{D}$
        \REQUIRE Defined goal state $\sim$ (CD, CM, ST)
        \WHILE{not done}
            \STATE Collect samples $\mathcal{D}^\pi$ on $N$ tasks
            \STATE Push samples to buffer $\mathcal{D}^\pi \rightarrow \mathcal{B}$
            \STATE \texttt{/* Hierarchical Training */}
            \FOR{minibatch $b$ in $\mathcal{B}$}
            \STATE Evaluate $\mathcal{L}_{HL}, \mathcal{L}_{IL}$ and gradients using equations \eqref{eqn:HL_loss} and \eqref{eqn:IL_loss}
                \STATE Update: Adam($\mathcal{L}_{HL}, \mathcal{L}_{IL}$) $\rightarrow$ $\phi, \psi, \theta_V$
            \ENDFOR
            \STATE \texttt{/* Policy Training */}
            \FOR{$K$ in PPO $K$-Epochs}
            \STATE Evaluate $\mathcal{L}_{LL} \sim \mathcal{D}^\pi$ using \Cref{eqn:ppo_objective}
            \STATE Update: Adam($\mathcal{L}_{LL}$) $\rightarrow$ $\theta_\pi$
            \ENDFOR
        \ENDWHILE
    \end{algorithmic}
\end{algorithm}

\subsection*{Analysis} \label{section:theoretical}
Here, we provide a mathematical view of macro-action discovery. To understand the trajectory of the system, consider the state transition over a discrete time interval \([t, t + M]\). Denoting the transition function at time \(t\) as \(T_t(s_t, a_t) = s_{t+1}\), the trajectory from time \(t\) to \(t + M\) can be represented as a composite function of \(T_t(\cdot)\):
\begin{equation} \label{eqn:composite_function}
    \begin{aligned}
        s_{t}, s_{t+1}, \cdots s_{t+M} = (T_{t+M-1} \circ \cdots \circ T_{t+1} \circ T_t)(s_t, a_t, \ldots, a_{t+M-1})
    \end{aligned}
\end{equation}
where \(T_{t+M - 1} \circ \cdots \circ T_{t+1} \circ T_t\) denotes the composition of transition functions with the input of state-action pairs between time \(t\) and  \(t + M - 1\).

In a model-free setting, the objective is to determine the next action $a_{t+1}$ based on the current state $s_t$, not the whole trajectory. Thus, we define a function approximator for learning such composite functions as in \Cref{eqn:composite_function} with one single input:
\begin{equation} \label{eqn:composite_function_single_input}
    s_{t+M} = f(s_t, z_t|\psi)
\end{equation}
where the neural approximator $\psi$ has at least two components for learning the composite function $(T_{t+M - 1} \circ \cdots \circ T_{t+1} \circ T_t)$ with the decoder $\psi_{s}$, and the action summary, $z$, of $(a_t, \cdots, a_{t + M - 1})$ with the encoder $\psi_z$. This reflects the design intent of an intermediate layer with an encoder and decoder for learning the composite function and the macro-action, respectively.

To further investigate the mathematical definition of $z$, we define a simple deterministic linear system:
\begin{equation} \label{eqn:linear_system}
    s_{t+1} = s_t + B\pi(s_t) = T_t(s_t, \pi(s_t))
\end{equation}
where $s_t \subset \mathbb{R}^s$ is the state at time $t$, $B \subset \mathbb{R}^{s\times a}$ is for a linear transformation that maps action to state space, $\pi(s_t)\subset \mathbb{R}^a$ is the action from policy, and $T$ is the transition function. Then, the recursive definition of the goal state under this system is 
\begin{equation} \label{eqn:recursive_goal_state}
    s_{t + M} = s_t + B\left(\sum_{t=t}^{t+M-1} \pi(s_t)\right)
\end{equation}
In above formulation, it is represented by the sum of the initial state and the transformation of the sum of actions, i.e., the \textit{macro-actions}. Our insight in \Cref{def:macro-action} is to approximate cumulative action embeddings using a missing-information technique with neural approximators for nonlinear system dynamics.

\subsection*{Challenges and Solutions}
We additionally addressed two major challenges to enhance overall performance. First, we tackle the ``curse of hierarchy," which refers to the mismatch in focus across layers. As the policy gathers new trajectories to maximize rewards, higher-level layers must adapt to new incoming dynamical information. This issue, caused by transitional shifts in hierarchical RL \cite{pateria2021hierarchical}, is mitigated through (1) a data buffer that refreshes with recent trajectories, (2) independent roles for each layer by designing each layer to achieve a specific goal and preventing gradient overflow off layers, and (3) streamlined architectures to avoid model complexity with which the higher-level models easily track the traces of low-level policy.

\begin{figure}[t!]
    \centering
    \includegraphics[width=0.4\textwidth]{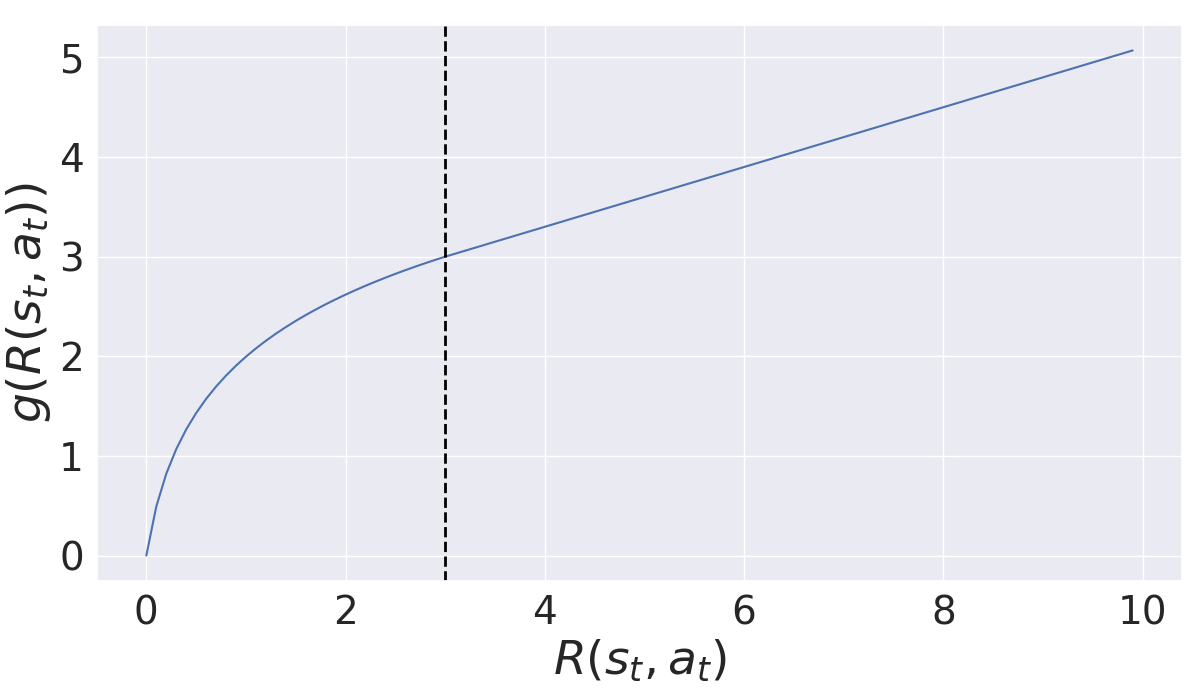}
    \caption{Reward shaping function with \(a=3\). Rewards below 3 is exponential, while the gradient is linear beyond \(a=3\).}
    \label{fig:reward_shape}
\end{figure}

Second, when learning multiple tasks, policies often prioritize easier tasks, leading to higher sample complexity and neglect of difficult tasks. To counter this, we adopt reward shaping with an asymptotic log function that boosts low rewards while maintaining the same increment rate beyond a specified point. This makes early rewards for difficult tasks are emphasized. For any non-negative rewards, the shaping function is:

\begin{equation} \label{eqn:reward_log_shape}
  g(x) = \begin{cases}
    \ln \left(\frac{(e^a - a)}{a}x + 1\right), & \text{if } x \leq a, \\
    \frac{a}{a^2 + 1}(x - a) + a, & \text{otherwise}.
  \end{cases}
\end{equation}
where \(a\) is a hyperparameter within the reward range. The reshaped rewards, \(g(R(s_t, a_t))\), are visualized with \(a=3\) in \Cref{fig:reward_shape}.

\section{Experiments} \label{section:Experiment}
We utilized MetaWorld \cite{yu2020meta} as our testbed. MetaWorld provides a structured environment for multi-task and meta-learning, designed to evaluate algorithms on their generalization capabilities. It improves upon earlier frameworks by systematically defining tasks with shared characteristics.

We compared our approach against the following baselines:
\begin{itemize}
    \item Learning Tasks via Subtask Decomposition \textbf{(SD\footnote{SDVT \cite{lee2023parameterizing} combines SD with \emph{virtual training} (VT), which adds computational complexity with minimal performance gains. Thus, we omit VT here for simplicity and it can be implemented per our method as well.})} \cite{lee2023parameterizing}
    \item Probabilistic Embeddings for Actor-Critic RL \textbf{(PEARL)} \cite{rakelly2019efficient}
\end{itemize}

\begin{figure}[t]
    \centering
    \begin{minipage}[b]{0.45\linewidth}
        \centering
        \includegraphics[width=\linewidth]{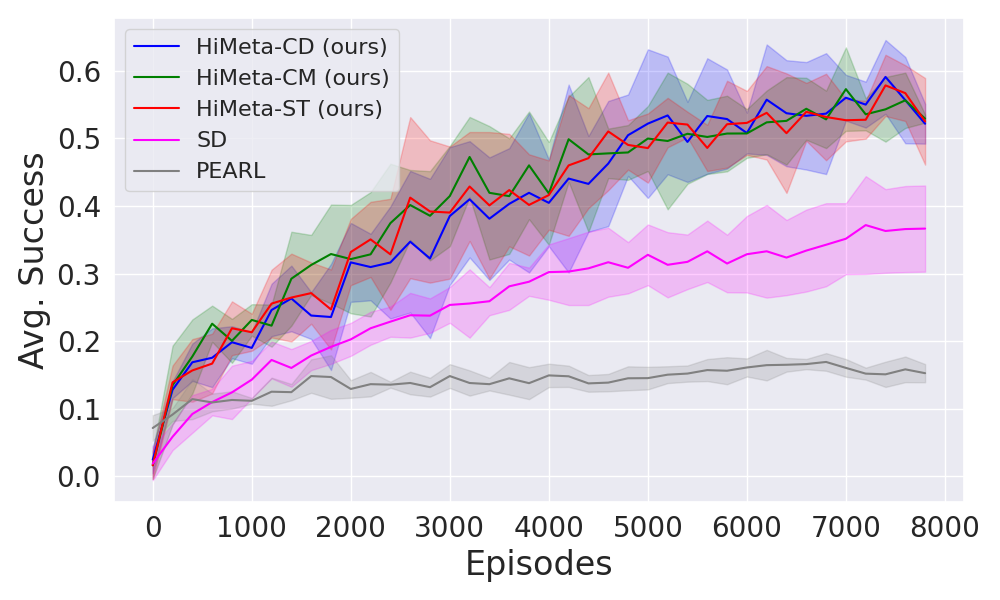}
        \caption*{(a) Success Metric}
        \label{fig:success}
    \end{minipage}
    \hfill
    \begin{minipage}[b]{0.45\linewidth}
        \centering
        \includegraphics[width=\linewidth]{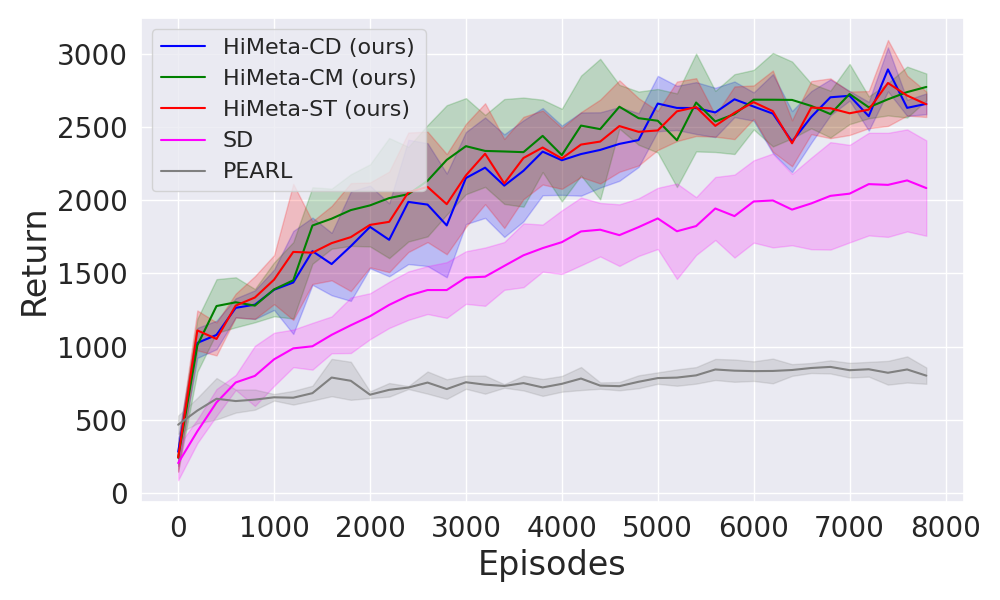}
        \caption*{(b) Reward Metric}
        \label{fig:reward}
    \end{minipage}
    
    \caption{MetaWorld ML10 learning curve is shown here. The mean and standard deviations are plotted along the average success and return metrics. (a) Cumulative success is plotted where success is set to 1 at the point the agent completes the task and afterward. (b) Cumulative rewards with $\gamma = 1$.}
    \label{fig:learning-curve}
\end{figure}

SD is selected for its recent superior performance in MetaWorld, and our method builds on PEARL by incorporating an intermediate-level layer and replacing the encoder with a recurrent unit. Detailed network parameters and hyperparameters are provided in the Appendix. 

\begin{table*}[t]
    \centering    
    \captionsetup{labelfont=bf}
    \begin{tabular}{@{\extracolsep{4pt}}c@{\extracolsep{8pt}} c@{\extracolsep{8pt}} c@{\extracolsep{8pt}} c@{\extracolsep{8pt}} c@{\extracolsep{8pt}} c@{\extracolsep{8pt}} c@{}}
        \toprule\toprule
        \textbf{ML10} & \multicolumn{2}{c}{\textbf{Avg. Success}} &\multicolumn{2}{c}{\textbf{Success}} & \multicolumn{2}{c}{\textbf{Returns}} \\ 
        \cmidrule(lr){2-3} \cmidrule(lr){4-5} \cmidrule(lr){6-7}
        \textbf{Methods} & \textbf{Train} & \textbf{Test} & \textbf{Train} & \textbf{Test} & \textbf{Train} & \textbf{Test} \\ 
        \midrule
        \textbf{HiMeta-CD} & $\mathbf{0.64\pm0.07}$ & $\mathbf{0.13\pm0.11}$ &$0.73\pm0.09$ & $\mathbf{0.30\pm0.10}$ & $3232\pm207$ & $\mathbf{501\pm361}$ \\ 
        \textbf{HiMeta-CM} & $0.64\pm0.10$ & $0.12\pm0.10$ & $\mathbf{0.73\pm0.07}$ & $0.29\pm0.12$ & $\mathbf{3306\pm309}$ & $485\pm344$ \\ 
        \textbf{HiMeta-ST} & $0.64\pm0.08$ & $0.12\pm0.08$ & $0.72\pm0.13$ & $0.29\pm0.11$ & $3237\pm296$ & $497\pm335$ \\ 
        \midrule
        \textbf{SD} \cite{lee2023parameterizing} & $0.42\pm0.09$ & $0.05\pm0.06$ &$0.60\pm0.04$ & $0.06\pm0.13$ & $2438\pm468$ & $422\pm290$ \\ 
        \textbf{PEARL} \cite{rakelly2019efficient} & | & | & $0.22\pm0.26$ & $0.01\pm0.02$ & $1034\pm1201$ & $340\pm494$ \\ 
        \bottomrule\bottomrule
    \end{tabular}
    \caption{All experiments were conducted under the same parameters settings for a detailed comparison. The table shows average success, previously used success metrics, and returns with means and standard deviations. The previous success rate, ranging from $0$ to $1$, is considered $1$ for the entire trajectory regardless of when the agent completed the task. In contrast, ours measures average success through each time step of the trajectory, highlighting our method is more effective since the increment from average success to success is comparatively smaller than the increment in the SD. Additionally, returns represent cumulative rewards with $\gamma=1$. These results suggest that our approach achieves significant performance gains by distributing information load and utilizing action-guided learning with \emph{macro-actions}. The specific performance for each task is attached in the \Cref{tab:task-specific-return} and \Cref{tab:task-specific-success}.}
    \label{tab:avg-result}
\end{table*}

\subsection*{Evaluation Metric}
MetaWorld introduced a success metric to complement reward-based evaluations. In our settings, we use an average episodic success metric that not only indicates task success but also measures how quickly the task is accomplished. Unlike \cite{yu2020meta}, which reports the maximum success rate achieved over the training process, or \cite{lee2023parameterizing}, which measures success at the final time step without considering the time taken, our metric provides a more nuanced view of performance. In particular, we measure how \emph{macro-actions} effectively guide policy via time-wise success considerations. However, we provide additional results for completeness using the previous success metric in \cite{lee2023parameterizing}.

\begin{table*}[h]
    \centering
    \captionsetup{justification=centering, labelfont=bf}
    \caption{Task-wise success rates with mean and standard deviations.}
    \label{tab:task-specific-success}
    \begin{tabular*}{\textwidth}{@{\extracolsep{\fill}} c | c c c c c @{}}
        \toprule\toprule
        \multicolumn{6}{c}{\textbf{Average Success (Train)}} \\ 
        \midrule
        \textbf{Task} & \textbf{HiMeta-CD} & \textbf{HiMeta-CM} & \textbf{HiMeta-ST} & \textbf{SD} & \textbf{PEARL} \\ 
        \midrule
        Reach & $\mathbf{0.73 \pm 0.11}$ & $0.64 \pm 0.3$ & $0.68 \pm 0.12$ & $0.34 \pm 0.13$ & $0.46 \pm 0.04$ \\ 
        Push & $\mathbf{0.73 \pm 0.04}$ & $0.69 \pm 0.15$ & $0.65 \pm 0.10$ & $0.07 \pm 0.11$ & $0.0 \pm 0.0$ \\ 
        Pick-place & $0.55 \pm 0.26$ & $\mathbf{0.70 \pm 0.14}$ & $0.55 \pm 0.26$ & $0.03 \pm 0.05$ & $0.0 \pm 0.0$ \\ 
        Door-open & $0.77 \pm 0.04$ & $0.72 \pm 0.06$ & $\mathbf{0.78 \pm 0.02}$ & $0.76 \pm 0.05$ & $0.25 \pm 0.03$ \\ 
        Drawer-close & $\mathbf{0.96 \pm 0.0}$ & $\mathbf{0.96 \pm 0.0}$ & $\mathbf{0.96 \pm 0.0}$ & $0.95 \pm 0.0$ & $0.74 \pm 0.07$ \\ 
        Button-press & $0.76 \pm 0.06$ & $0.71 \pm 0.08$ & $0.80 \pm 0.06$ & $\mathbf{0.88 \pm 0.01}$ & $0.35 \pm 0.03$ \\ 
        Peg-insert-side & $0.78 \pm 0.03$ & $\mathbf{0.80 \pm 0.03}$ & $0.79 \pm 0.03$ & $0.02 \pm 0.02$ & $0.0 \pm 0.0$ \\ 
        Window-open & $0.87 \pm 0.02$ & $\mathbf{0.88 \pm 0.01}$ & $0.87 \pm 0.02$ & $0.86 \pm 0.01$ & $0.43 \pm 0.1$ \\ 
        Sweep & $0.71 \pm 0.07$ & $\mathbf{0.75 \pm 0.04}$ & $0.72 \pm 0.08$ & $0.33 \pm 0.18$ & $0.0 \pm 0.0$ \\ 
        Basketball & $0.03 \pm 0.03$ & $0.05 \pm 0.06$ & $0.07 \pm 0.08$ & $\mathbf{0.30 \pm 0.30}$ & $0.0 \pm 0.0$ \\ 
        \midrule\midrule
        \multicolumn{6}{c}{\textbf{Average Success (Test)}} \\ 
        \midrule
        Drawer-open & $0.15 \pm 0.15$ & $0.17 \pm 0.17$ & $\mathbf{0.17 \pm 0.13}$ & $0.10 \pm 0.14$ & $0.02 \pm 0.01$ \\ 
        Door-close & $\mathbf{0.21 \pm 0.19}$ & $0.16 \pm 0.10$ & $0.19 \pm 0.09$ & $0.10 \pm 0.08$ & $0.03 \pm 0.02$ \\ 
        Shelf-place & $\mathbf{0.03 \pm 0.03}$ & $0.0 \pm 0.0$ & $0.01 \pm 0.02$ & $0.0 \pm 0.0$ & $0.0 \pm 0.0$ \\ 
        Sweep-into & $\mathbf{0.29 \pm 0.20}$ & $0.29 \pm 0.27$ & $0.27 \pm 0.20$ & $0.11 \pm 0.14$ & $0.0 \pm 0.0$ \\ 
        Lever-pull & $0.0 \pm 0.0$ & $\mathbf{0.02 \pm 0.03}$ & $0.0 \pm 0.0$ & $0.01 \pm 0.02$ & $0.0 \pm 0.01$ \\ 
        \bottomrule\bottomrule
    \end{tabular*}
\end{table*}

The evaluation is conducted using ML-10. This consists of 10 training tasks and 5 test tasks, expecting representations learned from training tasks transferred to performance in testing tasks. Sampling is based on 5 episodic samples per iteration for each task (a total of 50 trajectories). Given each episode lasting 500-time steps, $7.5$k iterations are taken for training.

\subsection*{Evaluation}
Evaluations were conducted with 5 seeds using the same hyperparameters and environmental settings for our method and baselines to allow a detailed comparison of convergence speed and final performance. The table presents average training and testing performance across seeds, measured by average success, success, and reward metrics. Results are summarized in \Cref{tab:avg-result}, with learning curves provided in \Cref{fig:learning-curve}.\footnote{One may notice a discrepancy between our experimental results and those in \cite{lee2023parameterizing}. This is attributed to the MetaWorld version, which underwent a significant update in 2023. Additional remarks are provided in the Appendix.} For completeness, task-specific learning results are included in \Cref{tab:task-specific-success} and \Cref{tab:task-specific-return}.

While our methods with three different goal states do not exhibit significant benefits over other definitions, they consistently outperform other approaches in terms of both success and return metrics under the same parametric and learning settings. The high variance in PEARL suggests that it can learn some tasks while struggling with others. \textbf{The latest state-of-the-art method, SD \cite{lee2023parameterizing}, which has outperformed all previous meta-learning algorithms (VariBad \cite{zintgraf2019varibad}, LDM \cite{lee2021improving}, $\text{RL}^2$ \cite{duan2016rl}, and PEARL \cite{rakelly2019efficient}), falls short compared to our approach.} Furthermore, the smaller gap of our method from average success to success in \Cref{tab:avg-result} indicates a faster task completion rate by the agent. In this context, our approach achieves state-of-the-art performance and efficiency in completing tasks.

\section{Related Workds} \label{section:relatedWorks}
Parameter-sharing explores \emph{``which model components to share"} across tasks. Based on metadata like task labels, models determine the extent of weight and bias sharing, making non-shared parameters task-specific. Fully shared parameter architectures are termed \emph{``hard-parameter sharing"}, while those with varying degrees of task-specific architectures are \emph{``soft-parameter sharing"}. This approach proves effective and straightforward in tasks with slight deviations across tasks \cite{pahari2022multi, wallingford2022task, sun2020adashare}.

Task grouping, in contrast, focuses on identifying tasks that can effectively share parameters by forming task groups. Unlike parameter-sharing, which assigns distinct parameters to each task, task-grouping simplifies model structures by grouping similar tasks together and treating them as a single task \cite{song2022efficient, fifty2021efficiently, standley2020tasks}.

While effective, these approaches have drawbacks, requiring domain-dependent expert knowledge and careful network separation. Consequently, relevant to ours, several works have attempted to discover shared structures across tasks, \(z\), and condition them into the input space, \(\pi(a|s,z)\). Instead of merely expanding the input space, this method enriches it with relevant task-wise information, merging operational spaces that share similarities and significantly reducing the information load on the policy. \cite{rakelly2019efficient} was a pioneering work that discovered latent task-wise information via posterior sampling by training alongside the value function. Additionally, \cite{zintgraf2019varibad, lee2023parameterizing} utilize recurrent encoder and dynamic decoder architectures for representation learning through state and reward dynamical predictions. This approach effectively discovers task-wise latent information, integrates similar tasks, and distinguishes different ones, facilitating multi-task and meta-learning.
 
\section{Conclusions} \label{section:concluding remarks}
Our approach introduces a novel hierarchical decomposition to alleviate the decision-making burden of policies in the context of \emph{hard-parameter sharing}. We propose \emph{macro-actions}, which provide compressed action information linking the current state to the handcrafted goal state. This is achieved by learning to produce both \emph{macro-actions} and a composite transition function using modified VAE. Our method demonstrates quicker task completion and outperforms previous baselines, which were known for their strong performance in training (multi-task) and test tasks (meta-learning). This advantage is attributed to our innovative approach to directly help policy with an additional action embedding module that prevents the loss of optimality in learned tasks while simultaneously learning others. 

\section{Discussions}
As we present \emph{novel} network architecture that successsfully distributes the decision burden, further exploration of its potential applications is necessary and promising. Future investigations may include offline multi-task or meta-learning \cite{mitchell2021offline, pong2022offline, cho2024out}, safe reinforcement learning settings \cite{achiam2017constrained, cho2024constrained}, and multi-modal learning scenarios \cite{yu2022multimodal, hu2021end}. In such settings, we anticipate that proposed \emph{macro-actions} will alleviate the problem-complexity due to difficult problem assumption, additional constraints, and additional prediction features in one place, respectively.

\section*{Acknowledgments}
We thank Dr. Sungkwang Mun for arranging and helping with the computational resources at the High-Performance Computing Center (HPC).

\bibliography{sample}

\appendix
\section*{Schematic Summary}
We employ three hierarchical layers, each with distinct roles to enhance the learning process. Specifically, we designed high, intermediate, and low-level layers for task inference, \emph{macro-action} discovery, and primitive actions, respectively, as illustrated in \Cref{fig:himeta_schematic}. 
\begin{itemize}
    \item \textbf{High-level layer}: The high-level layer is task-specific, analyzing the type of task representations, $y$, the agent is solving.
    \item \textbf{Intermediate-level layer}: The intermediate layer converts the representation $y$ into a task-agnostic \emph{macro-actions} $z$, providing high-level idea of cumulative actions embeddings.
    \item \textbf{Low-level layer}: Proximal policy optimization is used to reduce the sample complexity of meta-learning. Its input space is conditioned with each output of higher-level layers.
\end{itemize}

This training scheme simplifies policy decision-making using task representation with \emph{macro-actions} and encoding valuable transitional information from the current state to the goal state. This approach is beneficial for complex domains with multiple tasks, sparse rewards, and long horizons. Moreover, relying on the policy's exploration to define goal states in the IL layer reduces the decision burden by helping the policy retain previously learned tasks while acquiring new ones.

\section*{Implementation Details}
Here, we describe the details of network and training parameters for our method \Cref{tab:himeta-parameters} and baselines \Cref{tab:sd-parameters} with the source codes we used. 

\begin{table*}[h!]
    \centering
    \captionsetup{justification=centering, labelfont=bf}
    \caption{HiMeta Model Parameters}
    \label{tab:himeta-parameters}
    \begin{tabular*}{\textwidth}{@{\extracolsep{\fill}} l l l l l l l l @{}}
        \toprule\toprule
        \textbf{Category} & \textbf{Description} & \textbf{Notation} & \multicolumn{4}{c}{\textbf{Value}} \\ 
        \midrule
        \multirow{6}{*}{General} & Optimizer & - &  \multicolumn{4}{c}{Adam}  \\
        & Sample size & $\mathcal{D}$ &   \multicolumn{4}{c}{50 trajectories (5 for each task)}  \\
        & Buffer size & $\mathcal{B}$ &  \multicolumn{4}{c}{1000 trajectories}  \\
        & Discount factor & $\gamma$ &  \multicolumn{4}{c}{0.99}  \\
        & Loss Scalars & $\alpha_V, \alpha_{\epsilon}, \alpha_o$ & \multicolumn{4}{c}{20.0, 1.0, 1.0} \\ 
        & Gaussian Std Range & $\sigma_{min}, \sigma_{max}$ & \multicolumn{4}{c}{0.5, 1.5} \\
        \midrule
        \multirow{7}{*}{Policy parameters}
        & Algorithm & - & \multicolumn{4}{c}{PPO} \\
        & K-Epochs & - & \multicolumn{4}{c}{5} \\
        & Clip & $\epsilon$ & \multicolumn{4}{c}{0.2} \\
        & Entropy Scaler & $\alpha_2$ & \multicolumn{4}{c}{$1e-2$} \\
        & GAE & $\lambda$ & \multicolumn{4}{c}{0.90} \\
        & Gaussian Std Range & $\sigma_{min}, \sigma_{max}$ & \multicolumn{4}{c}{0.5, 1.5} \\
        & $a$ in Reward Log Shape & - & \multicolumn{4}{c}{3} \\

        \toprule
         & \textbf{Network} & \textbf{Notation} & \textbf{Type} & \textbf{Hidden Size} & \textbf{Act.} & \textbf{Dropout} & \textbf{LR} \\ 
        \midrule
        \multirow{5}{*}{High-level Layer} & Recurrent Unit & $\mathbf{\phi}_h$ & GRU & 1 layer, 256 & None & 0.0 & $5e-7$ \\
        & Categorical Unit & $\mathbf{\phi}_y$ & MLP & (512, 512) & ReLU & 0.7 & $5e-7$ \\
        & Value Function & $\mathbf{\theta}_V$ & MLP & (256, 256) & Tanh & 0.0 & $5e-7$ \\
        & State Embedding & - & MLP & (64) & Tanh & 0.0 & $5e-7$ \\
        & Action Embedding & - & MLP & (32) & Tanh & 0.0 & $5e-7$ \\
        & Reward Embedding & - & MLP & (16) & Tanh & 0.0 & $5e-7$ \\
        
        \midrule
        \multirow{3}{*}{Intermediate Layer} & Encoder & $\mathbf{\psi}_z$ & MLP & (128, 128, 64, 32) & ReLU & 0.0 & $5e-7$ \\
        & Decoder & $\mathbf{\psi}_s$ & MLP & (32, 64, 128, 128) & ReLU & 0.7 & $5e-7$ \\
        & State Embedding & - & MLP & (64) & Tanh & 0.0 & $5e-7$ \\
        
        \midrule
        \multirow{1}{*}{Low-level Layer} & Policy & $\mathbf{\theta}_\pi$ & MLP & (256, 256) & Tanh & 0.0 & $3e-7$ \\
        \bottomrule\bottomrule
    \end{tabular*}
\end{table*}

\begin{table*}[h]
    \centering
    \captionsetup{justification=centering, labelfont=bf}
    \caption{SD Model Parameters}
    \label{tab:sd-parameters}
    \begin{tabular*}{\textwidth}{@{\extracolsep{\fill}} l l l l l l l l @{}}
        \toprule\toprule
        \textbf{Category} & \textbf{Description} & \textbf{Notation} & \multicolumn{4}{c}{\textbf{Value}} \\ 
        \midrule
        \multirow{6}{*}{General} & Optimizer & - &  \multicolumn{4}{c}{Same as HiMeta | (used as in SD paper)}  \\
        & Sample size & $\mathcal{D}$ &   \multicolumn{4}{c}{Same as HiMeta}  \\
        & Buffer size & $\mathcal{B}$ &  \multicolumn{4}{c}{Same as HiMeta}  \\
        & Discount factor & $\gamma$ &  \multicolumn{4}{c}{Same as HiMeta | (used as in SD paper)}  \\
        & Gaussian Std Range & $\sigma_{min}, \sigma_{max}$ & \multicolumn{4}{c}{Same as HiMeta | (used as in SD paper)} \\
        & Loss Scalars & $\alpha_V, \alpha_{\epsilon}, \alpha_o$ & \multicolumn{4}{c}{10.0, 1.0, 1.0 | (used as in SD paper)} \\ 
        
        \midrule
        \multirow{7}{*}{Policy parameters}
        & Algorithm & - & \multicolumn{4}{c}{Same as HiMeta | (used as in SD paper)} \\
        & K-Epochs & - & \multicolumn{4}{c}{Same as HiMeta | (used as in SD paper)} \\
        & Clip & $\epsilon$ & \multicolumn{4}{c}{Same as HiMeta} \\
        & Entropy Scaler & $\alpha_2$ & \multicolumn{4}{c}{Same as HiMeta} \\
        & GAE & $\lambda$ & \multicolumn{4}{c}{Same as HiMeta | (used as in SD paper)} \\
        & Gaussian Std Range & $\sigma_{min}, \sigma_{max}$ & \multicolumn{4}{c}{Same as HiMeta | (used as in SD paper)} \\
        & $a$ in Reward Log Shape & - & \multicolumn{4}{c}{None} \\

        \toprule
         & \textbf{Network} & \textbf{Notation} & \textbf{Type} & \textbf{Hidden Size} & \textbf{Act.} & \textbf{Dropout} & \textbf{LR} \\ 
        \midrule
        \multirow{7}{*}{Representation Module} & Recurrent Unit & $\mathbf{\phi}_h$ & \multicolumn{5}{c}{Same as HiMeta | (used as in SD paper)} \\
        & State Decoder & $\mathbf{\phi}_s$ & MLP & (64, 64, 32) & ReLU & 0.7 & $5e-7$ \\
        & Reward Decoder & $\mathbf{\phi}_r$ & MLP & (64, 64, 32) & ReLU & 0.7 & $5e-7$ \\
        & Categorical Unit & $\mathbf{\phi}_y$ & MLP & (512, 512) & ReLU & 0.0 & $5e-7$\\
        & State Embedding & - & \multicolumn{5}{c}{Same as HiMeta | (used as in SD paper)} \\
        & Action Embedding & - & \multicolumn{5}{c}{Same as HiMeta} \\
        & Reward Embedding & - & \multicolumn{5}{c}{Same as HiMeta} \\
        
        \midrule
        \multirow{2}{*}{Policy} & Policy & $\mathbf{\theta}_\pi$ & \multicolumn{5}{c}{Same as HiMeta | (used as in SD paper)} \\
        & Value Function & $\mathbf{\theta}_V$ & \multicolumn{5}{c}{Same as HiMeta | (used as in SD paper)} \\
        \bottomrule\bottomrule
    \end{tabular*}
\end{table*}

\subsection{Experimental Remarks}
It is noteworthy that the experiments were conducted with identical parametric settings, as well as the same number of epochs and iterations per epoch for HiMeta, SD, and PEARL. The main discrepancy between our results and those presented in \cite{lee2023parameterizing} arises from the MetaWorld version, which received a substantial update in 2023, resulting in significant changes to certain tasks. Additionally, the PEARL results in SD are based on $1000$ epochs and $4000$ iterations, as they reported using the initial parameters provided by \emph{garage}\footnote{\url{https://github.com/rlworkgroup/garage/blob/master/src/garage/examples/torch/pearl_metaworld_ml10.py}}. This number of iterations is considerably larger than ours, which consisted of $40$ epochs and $200$ iterations. Therefore, the large variance in the PEARL results can be attributed to under-converged results and our commitment to using \emph{exactly} equivalent learning settings for a precise comparison of performance in terms of efficiency.
\begin{itemize}
    \item \textbf{SD}: https://github.com/suyoung-lee/SDVT
    \item \textbf{PEARL}: https://github.com/rlworkgroup/garage/pull/2287
\end{itemize}

\begin{table*}[h!]
    \centering
    \captionsetup{labelfont=bf}
    \caption{Task-wise returns with mean and standard deviations. Here, SD achieves a higher return during testing despite a lower success metric, consistent with the intuition behind the success metric proposed by \cite{yu2020meta}. This behavior arises because our method ceases activity after task success, whereas SD continues to act throughout the trajectory without succeeding.}
    \label{tab:task-specific-return}
    \begin{tabular*}{\textwidth}{@{\extracolsep{\fill}} c | c c c c c @{}}
        \toprule\toprule
        \multicolumn{6}{c}{\textbf{Return (Train)}} \\ 
        \midrule
        \textbf{Task} & \textbf{HiMeta-CD} & \textbf{HiMeta-CM} & \textbf{HiMeta-ST} & \textbf{SD} & \textbf{PEARL} \\ 
        \midrule
        Reach & $\mathbf{4502 \pm 165}$ & $4400 \pm 432$ & $4315 \pm 364$ & $3528 \pm 704$ & $2810 \pm 240$ \\ 
        Push & $\mathbf{3688 \pm 127}$ & $3446 \pm 841$ & $3149 \pm 606$ & $1035 \pm 198$ & $17 \pm 4$ \\ 
        Pick-place & $1434 \pm 450$ & $\mathbf{2472 \pm 410}$ & $1964 \pm 833$ & $198 \pm 217$ & $5 \pm 1$ \\ 
        Door-open & $4060 \pm 154$ & $3849 \pm 387$ & $\mathbf{4113 \pm 143}$ & $4076 \pm 195$ & $1303 \pm 106$ \\ 
        Drawer-close & $\mathbf{4786 \pm 39}$ & $4695 \pm 165$ & $4771 \pm 33$ & $4743 \pm 41$ & $3358 \pm 312$ \\ 
        Button-press & $3414 \pm 155$ & $3312 \pm 295$ & $\mathbf{3505 \pm 143}$ & $3089 \pm 118$ & $1130 \pm 109$ \\ 
        Peg-insert-side & $3822 \pm 81$ & $\mathbf{4020 \pm 82}$ & $3902 \pm 139$ & $340 \pm 366$ & $6 \pm 1$ \\ 
        Window-open & $3989 \pm 169$ & $4063 \pm 25$ & $3943 \pm 85$ & $\mathbf{4148 \pm 160}$ & $1661 \pm 187$ \\ 
        Sweep & $3758 \pm 369$ & $\mathbf{3964 \pm 210}$ & $3811 \pm 387$ & $1925 \pm 999$ & $42 \pm 6$ \\ 
        Basketball & $505 \pm 161$ & $\mathbf{588 \pm 131}$ & $583 \pm 186$ & $1416 \pm 1274$ & $7 \pm 1$ \\ 
        \midrule\midrule
        \multicolumn{6}{c}{\textbf{Return (Test)}} \\ 
        \midrule
        Drawer-open & $1592 \pm 405$ & $1557 \pm 419$ & $1546 \pm 338$ & $\mathbf{2325 \pm 224}$ & $1298 \pm 177$ \\ 
        Door-close & $523 \pm 198$ & $502 \pm 124$ & $525 \pm 125$ & $\mathbf{571 \pm 222}$ & $152 \pm 101$ \\ 
        Shelf-place & $\mathbf{267 \pm 279}$ & $97 \pm 70$ & $236 \pm 214$ & $153 \pm 164$ & $0.0 \pm 0.0$ \\ 
        Sweep-into & $994 \pm 873$ & $\mathbf{1124 \pm 1074}$ & $996 \pm 912$ & $662 \pm 740$ & $31 \pm 8$ \\ 
        Lever-pull & $219 \pm 92$ & $220 \pm 109$ & $229 \pm 88$ & $\mathbf{300 \pm 36}$ & $218 \pm 33$ \\ 
        \bottomrule\bottomrule
    \end{tabular*}
\end{table*}

\end{document}